# MGML: A Plug-and-Play Meta-Guided Multi-Modal Learning Framework for Incomplete Multimodal Brain Tumor Segmentation

Yulong Zou[1,2], Bo Liu[1], Cun-Jing Zheng[3], Yuan-ming Geng[4], Siyue Li[5], Qiankun Zuo[6,*], Shuihua Wang[7], Yudong Zhang[8], Jin Hong[2,*]

1. School of Mathematics and Computer Sciences, Nanchang University, 330031, China
2. School of Information Engineering, Nanchang University, Nanchang, 330031, China
3. Department of Radiology, Sun Yat-Sen Memorial Hospital, Sun Yat-Sen University, Guangzhou, 510120, China
4. Department of Stomatology, Zhujiang Hospital, Southern Medical University, 510282, Guangzhou, China
5. Department of Radiological Sciences, University of California Los Angeles, Los Angeles, CA, United States
6. School of Information Engineering, Hubei University of Economics, Wuhan 430205, Hubei China
7. Department of Biological Sciences, School of Science, Xi'an Jiaotong Liverpool University, Suzhou, 215123, China
8. School of Computer Science and Engineering, Southeast University, Nanjing, 210096, China

E-mail: 9109223090@email.ncu.edu.cn; liuboncu@email.ncu.edu.cn; zhengcj@mail2.sysu.edu.cn; gym@smu.edu.cn; siyueli@mednet.ucla.edu; qk.zuo@hbue.edu.cn; Shuihua.Wang@xjtlu.edu.cn; yudongzhang@seu.edu.cn; hongjin@ncu.edu.cn;

* Correspondence should be addressed to Qiankun Zuo & Jin Hong

**Abstract**: Leveraging multimodal information from Magnetic Resonance Imaging (MRI) plays a vital role in lesion segmentation, especially for brain tumors. However, in clinical practice, multimodal MRI data are often incomplete, making it challenging to fully utilize the available information. Therefore, maximizing the utilization of this incomplete multimodal information presents a crucial research challenge. We present a novel meta-guided multi-modal learning (MGML) framework that comprises two components: meta-parameterized adaptive modality fusion and consistency regularization module. The meta-parameterized adaptive modality fusion (Meta-AMF) enables the model to effectively integrate information from multiple modalities under varying input conditions. By generating adaptive soft-label supervision signals based on the available modalities, Meta-AMF explicitly promotes more coherent multimodal fusion. In addition, the consistency regularization module enhances segmentation performance and implicitly reinforces the robustness and generalization of the overall framework. Notably, our approach does not alter the original model architecture and can be conveniently integrated into the training pipeline for end-to-end model optimization. We conducted extensive experiments on the public BraTS2020 and BraTS2023 datasets. Compared to multiple state-of-the-art methods from previous years, our method achieved superior performance. On BraTS2020, for the average Dice scores across fifteen missing modality combinations, building upon the baseline, our method obtained scores of 87.55, 79.36, and 62.67 for the whole tumor (WT), the tumor core (TC), and the enhancing tumor (ET), respectively. We have made our source code publicly available at https://github.com/worldlikerr/MGML.

**Keywords**：Multimodal Brain Tumor Segmentation; Incomplete Multimodal Learning; Meta-parameterization; Self-supervised Learning; Knowledge Distillation

# 1. Introduction

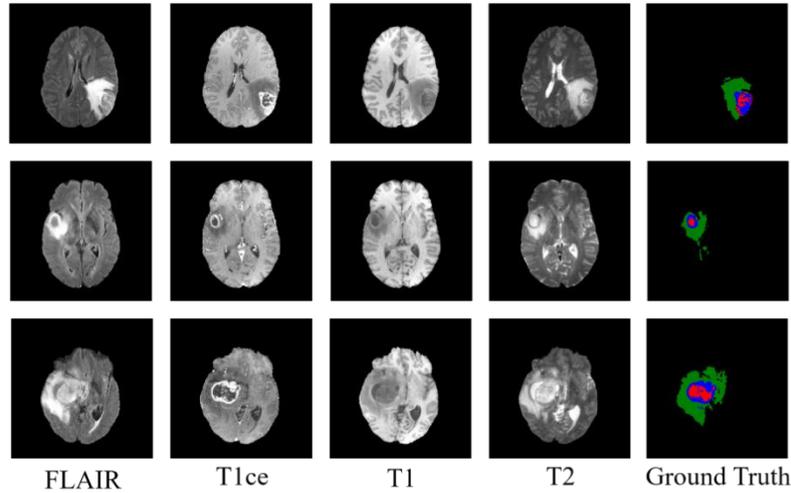

Figure 1. Visualization of how different modalities contribute to the segmentation of various brain tumor regions. The left panel shows the four modalities (T1, T1ce, T2, FLAIR). The right panel displays the final segmentation results, where green, blue, and red highlight the whole tumor (WT), the tumor core (TC), and the enhancing tumor (ET), respectively.

With the development of computer vision in the medical field, MRI has become increasingly significant for medical image segmentation. Its multiple modalities provide distinct contrasts and clues for different histological components of tumors, particularly in the domain of brain tumor segmentation. As shown in Figure 1, different modalities have varying sensitivities to different brain tumor regions. The BraTS benchmark [1] standardizes brain tumor segmentation into three regions: the whole tumor (WT), the tumor core (TC), and the enhancing tumor (ET), based on multimodal MRI inputs (T1, T1c, T2, FLAIR). Specifically, T1c shows enhancing components and is the primary source for segmenting ET. T2/FLAIR is most sensitive to edema/infiltration, making it key for delineating the WT region. T1, where tumors often appear as low-signal regions, provides contrast with gray and white matter for structural localization. While leveraging multimodal information can significantly improve segmentation accuracy, missing MRI modalities are a common occurrence in clinical practice due to resource limitations and patient conditions. Therefore, achieving advanced segmentation performance despite missing modality inputs is a valuable research challenge.

Common approaches for handling missing modalities in recent years include missing modality imputation, optimization of multimodal fusion architectures, and cross-modal knowledge distillation.

**Missing Modality Imputation.** This approach often uses generative adversarial networks (GANs) [2], [3] and diffusion models [4] to reconstruct missing modalities from existing information. For instance, Hong Liu et al. [5] proposed a novel two-stage framework for missing modality imputation, while Heran Yang et al. [6] introduced a joint feature- and image-level modality completion method. However, these methods often struggle to balance the plausibility of the generated modalities with low computational costs.

**Multimodal Fusion Architecture Optimization.** This involves modifying model architectures to better handle multimodal data. For example, Yuhang Ding et al. [7] proposed a region-aware hierarchical pyramid structure (RFNet) that achieved a significant performance boost over SOTA methods at the time. Building on RFNet, Yao Zhang et al. [8] proposed mmFormer, which introduced a Transformer architecture to enhance the model's global awareness and modality alignment, overcoming the limited receptive fields of pure CNNs. Considering the quadratic complexity of Transformers, LS3M [9] introduced a State Space Model (SSM) into a CNN-based encoder-decoder framework to efficiently model both intra- and inter-modality long-range dependencies. While these methods improve fusion capabilities, they require changes to the original network architecture and introduce new parameters, leading to a greater computational burden.

**Cross-modal Knowledge Distillation.** Methods like those in [10] and [11] use cross-modal knowledge distillation to transfer knowledge between strong and weak modalities, promoting information fusion. However, they are unable to adaptively mitigate the negative impacts of information transfer from different modalities.

To overcome the limitations of previous approaches, we present a novel meta-guided multi-modal learning (MGML) framework, which consists of two key components: meta-parameterized adaptive modality fusion (Meta-AMF) and consistency regularization module, as shown in Figure 2. The Meta-AMF module, our primary contribution, is composed of a meta-parameterized network (MetaNetwork) and an adaptive modality fusion (AMF) component. During training, the MetaNetwork receives segmentation predictions from the shared decoder of each modality and adaptively generates meta-parameters tailored for the current multi-modal input data. These meta-parameters formulate a more favorable fusion strategy for the specific input. The AMF module then uses these meta-parameters and the prediction results from each modality to dynamically adjust the degree of high-confidence and conservative fusion, as well as the importance ratio between them. This process generates a meta-parameter-guided soft label that acts as an additional supervision signal, explicitly promoting the fusion of multimodal information. As our secondary contribution, we adopt a teacher–student paradigm. It treats the model's states under complete and incomplete modality inputs as teacher-like and student-like models, respectively, without relying on an additional powerful teacher network. By computing a loss between these two states, we introduce an auxiliary regularization term in the total loss. This term improves model robustness and generalization, while implicitly promoting multimodal information fusion.

Unlike existing meta-learning-based distillation methods such as MetaKD[12], which focus on transferring knowledge between modality-specific networks. The resulting MGML keeps the original model architecture unchanged and can be seamlessly integrated into the training pipeline for end-to-end learning. This meta-guided adaptive fusion enables implicit self-distillation within a single model, improving robustness to missing modalities while maintaining architectural simplicity. In summary, our contributions are as follows:

(i) We propose a novel meta-guided multi-modal learning (MGML) framework for incomplete multimodal brain tumor segmentation, which can be seamlessly integrated into existing models without altering their original architecture.

(ii) Within MGML, we design a meta-parameterized adaptive modality fusion (Meta-AMF) module that adaptively generates meta-parameters to guide a more favorable fusion strategy, explicitly enhancing the integration of multimodal information.

(iii) We further introduce consistency regularization module that leverages a teacher–student paradigm to provide auxiliary regularization, thereby improving model robustness and implicitly promoting multimodal fusion.

(iv) Extensive experiments on the BraTS2020 and BraTS2023 datasets demonstrate the effectiveness and superiority of MGML, validating its ability to enhance the performance of state-of-the-art methods under various missing-modality scenarios.

## 2. Related Work

**Medical Image Segmentation.** Traditional medical image segmentation tasks rely on encoder-decoder network architectures, often using convolutional neural networks (CNNs) as their foundational building blocks [13], [14]. This approach has achieved strong segmentation performance while maintaining a lightweight model footprint. The introduction of nnU-Net [15], which demonstrates state-of-the-art performance on a wide range of datasets, has provided an out-of-the-box solution, allowing various subfields of image segmentation to leverage this advanced technology [16]. Originally developed for natural language processing (NLP) tasks, Transformers [17] have also been widely adopted in image segmentation [18], [19]. In certain segmentation tasks, Transformer-based models have even outperformed pure CNN-based architectures.

Yet these methods mentioned above often struggle to accurately simulate missing modalities during training and fail to fully leverage the information from multiple modalities. Consequently, in clinical practice where multi-modal data is often incomplete, these models cannot achieve the same level of advanced performance seen with complete datasets.

**Incomplete Multimodal Learning.** Multimodal fusion often provides richer information compared to single-modality data. However, in clinical practice, it's common for the number of modalities for a given segmentation task to be incomplete, leading to a certain amount of missing data. Previous research has focused on two main approaches: missing modality imputation and model architecture optimization to better leverage multimodal information.

For the former, the development of generative adversarial networks (GANs) [20] has advanced the field of missing modality imputation [2], [3]. These models can generate realistic missing modalities based on existing ones, which can then be used in downstream multimodal information fusion tasks. More recently, with the advent of modern diffusion models [21], they have also been applied to modality imputation [4]. Other works [5], [6] are exploring their own unique methods for modality completion. For the latter, RFNet [7] proposed a hierarchical pyramid structure that, through its region-aware capability for brain tumors, achieved superior performance in incomplete multimodal brain tumor segmentation. Building on RFNet, both mmFormer [8] and M2FTrans [22] introduced the Transformer architecture to enhance the model's global awareness and modality alignment capabilities, thereby overcoming the local receptive field limitations of pure CNN architectures. Following mmFormer, IM-Fuse [23] incorporated the SSM-based Mamba module [24] into a fused CNN-Transformer architecture, achieving further performance improvements. Given the additional computational burden caused by the quadratic complexity of the Transformer, LS3M [9] introduced an improved SSM module into an encoder-decoder framework. This approach efficiently models both intra-modality and inter-modality long-range dependencies.

Nevertheless, these methods either require the introduction of complex generative adversarial networks or diffusion models, or they necessitate constant modifications to the existing network architecture by adding new modules. This continuous addition of complexity has led to an increasingly severe computational burden during the training process.

**Knowledge Distillation.** Knowledge distillation (KD) is an effective model optimization strategy, originally proposed in [25], where a lightweight student network is trained by leveraging the rich knowledge of a powerful teacher network. Initial KD techniques were refined and applied to simpler tasks [26], [27]. As research has advanced, more sophisticated methods for knowledge transfer have emerged.

In [28], a feature-based distillation approach was used to transfer structured knowledge from the teacher to the student network through a pixel-wise distillation technique. For multimodal tasks, [10] proposed a cross-modal knowledge distillation method where the teacher and student networks process different modalities, with knowledge being transferred between them to improve segmentation performance. A cross-modal self-distillation method was introduced in [11] to enable a segmentation network for a weaker modality to learn from a stronger one, thereby enhancing the quality of information after multimodal fusion. On the other hand, such strategies require modeling both the teacher and student networks. Even with a relatively lightweight student model, this approach introduces an extremely large number of models and parameters, which can be computationally prohibitive for the task of missing multimodal brain tumor segmentation.

Among meta-learning–based distillation approaches, MetaKD [12] introduced a meta-learned modality-weighted knowledge distillation strategy that estimates modality importance and transfers knowledge between separate modality-specific networks. While effective, such frameworks rely on cross-modal teacher–student architectures and focus on reweighting distillation terms rather than improving the fusion process itself.

In contrast, our proposed MGML framework adopts a fundamentally different perspective: instead of using meta-learning to determine distillation weights, we employ a meta-parameterized controller to directly guide the fusion of multi-modal predictions within a single model. This design removes the need for additional teacher networks, enables plug-and-play integration with existing architectures, and facilitates end-to-end optimization of both fusion behavior and segmentation performance.

## 3. Method

In this section, we introduce our MGML framework for incomplete multimodal brain tumor segmentation. It consists of two modules: Meta-AMF for explicit fusion and a consistency regularization module for implicit fusion.

## 3.1. Overall Architecture of the X+MGML Framework

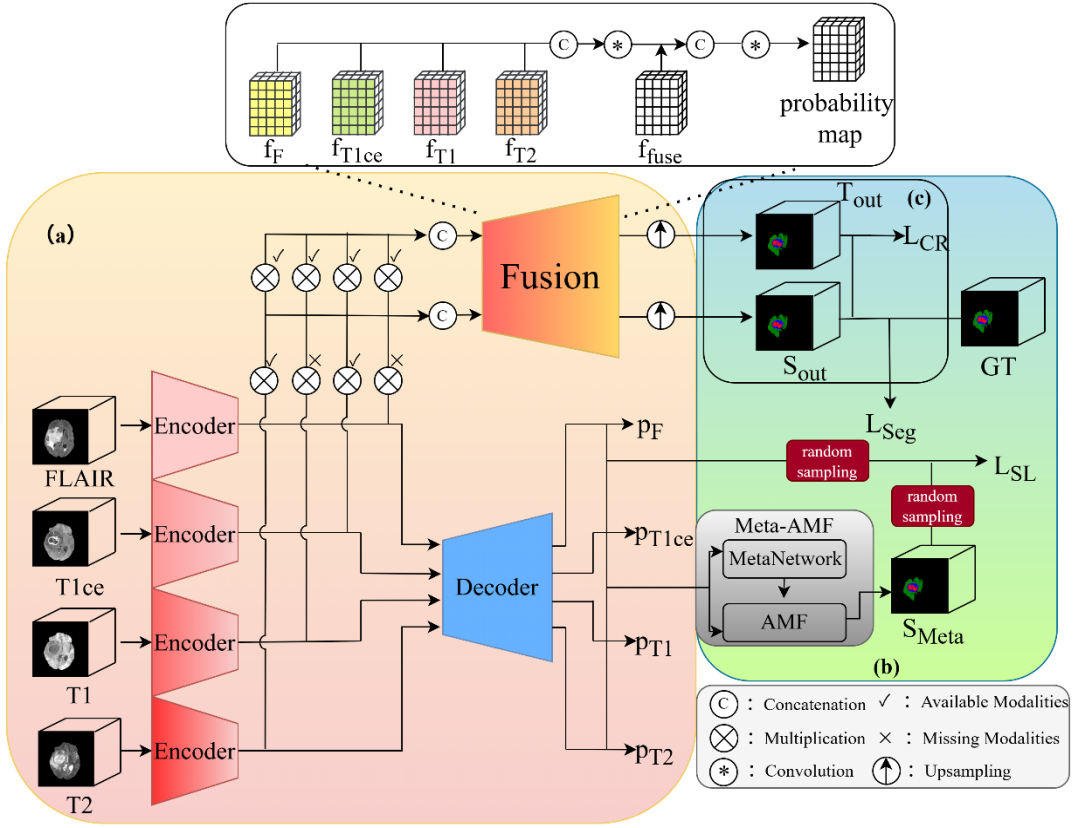

Figure 2. The overall architecture and design details of the X+MGML framework. (i) Subfigure (a) illustrates the existing incomplete brain tumor segmentation method X.(ii) Subfigure (b) presents the proposed MGML framework, where (c) highlights the teacher-student paradigm with consistency regularization module.

As depicted in Figure 2, our model is built upon the 3D U-Net architecture [29]. All four MRI modalities are simultaneously provided as inputs to the network. It follows a modality-specific segmentation design, where each modality is processed by an independent encoder and a shared decoder maps them into a common latent space. The multi-modal fusion strategy from [7], [8] (denoted as the Fusion module in Figure 2) is adopted as the baseline fusion scheme to produce the final segmentation outputs.

On top of this baseline, we introduce our meta-guided multi-modal learning (MGML) framework. Specifically, the meta-parameterized adaptive modality fusion (Meta-AMF) module receives modality-specific predictions from the shared decoder and adaptively generates soft supervisory signals through dynamic fusion. These soft labels are then used in a self-distillation loss to refine each modality's prediction. Furthermore, we incorporate a teacher-student paradigm in which the full-modality prediction (teacher-like) supervises the incomplete-modality prediction (student-like) through a consistency-regularized loss. This provides an additional self-supervised signal that enhances robustness and promotes more reliable multi-modal fusion. Importantly, the proposed MGML framework does not modify the baseline network architecture. As a lightweight and plug-and-play component, it can be seamlessly integrated into existing multi-modal segmentation frameworks for end-to-end training.

## 3.2. Meta-Parameterized Adaptive Modality Fusion for Soft Label Distillation

**Comparison with MetaKD and Design Motivation.** MetaKD[12] addresses the missing-modality problem by performing cross-modal knowledge distillation between multiple modality-specific subnetworks. It leverages a meta-learner to assign importance weights to each modality, and transfers knowledge from high-accuracy modalities to lower-accuracy ones through weighted distillation losses. While this design effectively models modality importance, it requires a set of independent teacher–student networks and focuses on optimizing distillation weights rather than improving feature fusion itself. As a result, its scalability and integration into arbitrary segmentation architectures are limited.

In contrast, our Meta-Guided Multi-Modal Learning (MGML) framework introduces a Meta-AMF module that redefines the role of meta-learning in multimodal fusion. Instead of supervising cross-network knowledge transfer, MGML employs a meta-parameterized controller within a single unified model to directly guide adaptive fusion of modality-specific predictions. The MetaNetwork learns task-adaptive fusion coefficients conditioned on the current input and prediction confidence, dynamically balancing Smooth Max (aggressive) and Smooth Min (conservative) fusion behaviors to produce high-quality soft labels for distillation.

This re-formulation provides several practical advantages: (i)Plug-and-play integration – MGML can be seamlessly embedded into existing multimodal segmentation architectures (e.g., RFNet, mmFormer, IM-Fuse) without modifying their backbone design; (ii)Self-contained distillation – by performing implicit teacher–student alignment within one model, MGML removes the need for external teacher networks; (iii) Adaptive optimization – the meta-parameterized fusion allows the network to self-adjust under diverse missing-modality conditions.

The overall pipeline of Meta-AMF is illustrated in Fig. 2, where the MetaNetwork and Adaptive Modality Fusion components jointly generate input-specific soft labels that stabilize the learning process and improve segmentation accuracy.

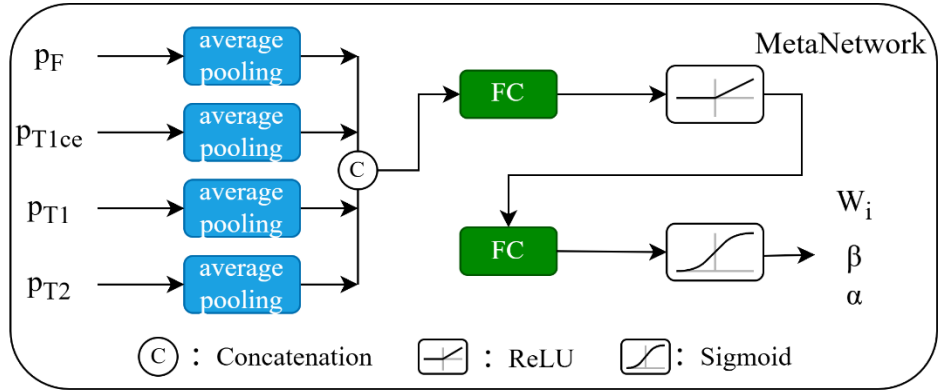

Figure 3. A schematic diagram of our proposed MetaNetwork module. Where $w_f$, $\beta$ and $\alpha$ are the meta-parameters generated by the MetaNetwork.

**Meta-Parameterized Network.** Meta-learning is a machine learning paradigm where a model is trained on a variety of different but related tasks. This allows it to learn a general form of "meta-knowledge" that can quickly adapt to new tasks. Building on the work of [30] et al., the concept of meta-learning has been widely applied in deep learning.

Different from previous meta-learning–based distillation frameworks, our MetaNetwork directly generates task-adaptive parameters to guide the fusion process, avoiding the need for separate teacher models. Based on a global perception of the current multi-modal image predictions, it dynamically outputs a series of parameters. These parameters guide the adaptive modality fusion (AMF) module to perform the most suitable modality fusion. The final soft labels output by the Meta-AMF and the $\mathcal{L}_{SL}$ from each modality are then used in the back propagation process to optimize the model parameters.

As shown in Figure 2, the Meta-AMF receives predicted probability maps for each modality from the shared decoder. Subsequently, as depicted in Figure 3, the MetaNetwork processes the input modalities by applying global average pooling to each and then concatenating them. This process is formulated as:

$$z = \text{GAP}(P_1, P_2, \ldots, P_M) \in R^{B \times (M \cdot C)}, \quad (1)$$

where $P_i$ represents the predicted probability map for the $i$-th modality, and GAP denotes global average pooling.

Within the MetaNetwork, we introduce a small network consisting of two fully connected layers and two activation functions, as shown in Figure 3. The processed features are then mapped to output the final meta-parameters $T_1, T_2, w_f, \beta, \alpha$, expressed as:

$$[T_1, T_2, w_f, \beta, \alpha] = f_{\text{meta}}(z), \quad (2)$$

where the mapping for parameter is $\beta = \sigma(p_\beta) \cdot (\beta_{\max} - \beta_{\min}) + \beta_{\min}$ (the mappings for the other meta-parameters are identical).

Among the output meta-parameters: $T_1, T_2$ acts as an internal learning signal, implicitly guiding the MetaNetwork on how to evaluate the confidence of multi-modal fusion for the current data, thereby influencing the generation of $w_f, \beta$ and $\alpha$. $w_f$ determines the weights of the high-confidence fusion (Smooth Max) and the conservative fusion (Smooth Min) in the final fused result. This is the core meta-parameter that reflects the overall confidence assessment. $\beta$ is used to control the sharpness of the high-confidence fusion (Smooth Max), i.e., its approximation to the hard Max operation. $\alpha$ is used to control the sharpness of the conservative fusion (Smooth Min), i.e., its approximation to the hard Min operation.

In summary, based on the current input features, the MetaNetwork obtains an overview of the overall quality and confidence of the current batch of data. By backpropagating optimization parameters through the final soft-label loss, the network learns to infer the overall confidence state of a given sample from global statistical information. This, in turn, allows it to output meta-parameters specifically tailored for the current modality combination.

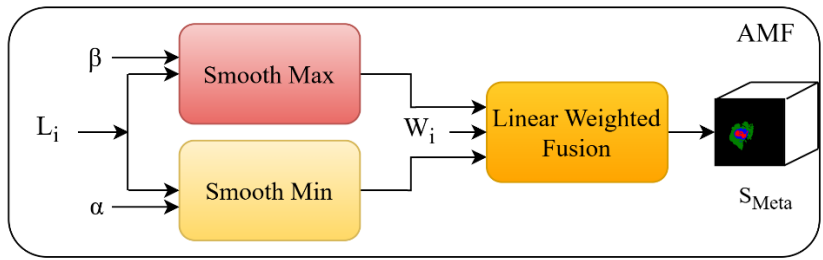

Figure 4. A schematic diagram of our proposed adaptive modality fusion (AMF) module. Where $w_f$, $\beta$ and $\alpha$ are the meta-parameters generated by the MetaNetwork. $L_i$ is the logits output of the $i$-th modality.

**Adaptive Modality Fusion Module.** Under the macro-control of the meta-parameters generated by the MetaNetwork, the AMF serves as the core fusion function for the modality predictions. It accepts the original, pre-softmax logits from all modalities and, through its internal dynamic fusion mechanism, produces a specific soft label tailored for the current multi-modal input data.

As shown in Figure 4, the AMF first receives the raw logits from each modality and performs high-confidence fusion via the Smooth Max module, as follows:

$$H = \frac{1}{\beta} \log \left( \sum_{i=1}^{M} \exp(\beta L_i) \right), \tag{3}$$

where $L_i$ is the logits output of the $i$-th modality, $M$ is the number of modalities, and $\beta$ is the meta−parameter output from the MetaNetwork. The parameter $\beta$ contains knowledge learned during training, guiding the AMF to decide, for each location of the current multi-modal input, whether to aggressively select the prediction from the strong, high-confidence modality or to use a softer approach by performing a weighted sum of predictions across all modalities.

Simultaneously, a conservative fusion is performed by the Smooth Min module, as follows:

$$C = -\frac{1}{\alpha} \log \sum_{i=1}^{M} \exp(-\alpha L_i), \tag{4}$$

Similarly, guided by the meta-parameter $\alpha$ output by the MetaNetwork based on its accumulated training experience, the AMF determines for each location whether to aggressively select the minimum value or to perform a soft weighted sum.

Through the operations above, we obtain the high-confidence fusion result $H$ and the conservative fusion result $C$. These results, along with the meta-parameter $w_f$ from the MetaNetwork, are then fed into the Linear Weighted Fusion module, as shown in Figure 3. This module achieves a dynamic weighted combination of the multi-modal fusion information, as follows:

$$S = w_f \cdot H + (1 - w_f) \cdot C, \tag{5}$$

This formula uses $w_f$ to linearly interpolate between the outputs of the two dynamic fusion strategies, thereby determining the form of the final dynamically weighted result. In the MetaNetwork, $w_f$ is mapped to a value between 0 and 1. If $w_f$ approaches 1, it indicates that the MetaNetwork is optimistic about the current multi−modal data prediction and tends to adopt an aggressive, high−confidence fusion strategy. If $w_f$ is close to 0, it suggests that the prediction is less certain, and the MetaNetwork tends to adopt a pessimistic and conservative fusion strategy. This dynamic weighting method of the AMF outputs a superior soft label for the current multi-modal input data.

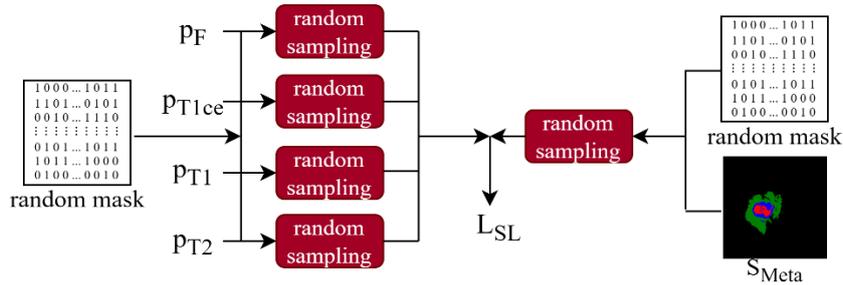

Figure 5. A diagram illustrating the process of applying random masking to predictions from each modality and to the soft labels, followed by the calculation of the loss.

**Confidence-Guided Random Masking.** Through the preceding operations, we obtain a specific soft label for the current multi-modal data input. Inspired by [31] and other mask-based pre-training methods, which have shown that random masking can improve model robustness without degrading performance, we adopt a similar random masking strategy. Specifically, we apply a randomly generated mask to both the soft labels ($S_{meta}$) produced by the Meta-AMF module and the logits from each modality. Unlike MAE-based approaches, our random mask sets values to zero at the same spatial positions across all modalities, effectively discarding potentially inaccurate predictions. In addition, we incorporate local confidence information when generating the random mask, giving higher masking probability to low-confidence voxels while preserving reliable regions for effective distillation. As show in Figure 5, we apply the random masking process to both the soft labels ($S_{meta}$) and the logits output from each modality, as shown below:

$$\widetilde{P}_i = \text{Mask}\left(P_i, \mathcal{M}_{w_f}\right), \tag{6}$$

Here, $P_i$ denotes the original logits output from the $i$-th modality, and $\mathcal{M}_{w_f}$ represents the mask matrix dynamically generated according to the overall confidence $w_f$ of the current batch. Analogously,

$$\tilde{S}_{meta} = \text{Mask}\left(S_{meta}, \mathcal{M}_{w_f}\right)$$

**Soft Label Distillation Loss.** We employ a Weighted Cross-Entropy (WCE) loss to compute the soft-label distillation loss between the randomly logits $\widetilde{P}_i$ and soft labels $\tilde{S}_{meta}$, formulated as:

$$\mathcal{L}_{\text{SL}} = \frac{1}{M}\sum_{i=1}^{M} \text{WCE}\left(\widetilde{P}_i, \tilde{S}_{meta}\right), \tag{7}$$

where $WCE(\cdot,\cdot)$ represents the Weighted Cross-Entropy loss, which is calculated as follows:

$$\text{WCE}\left(\widetilde{P}_i, \tilde{S}_{meta}\right) = -\frac{1}{B}\sum_{b=1}^{B}\sum_{c=1}^{C} w_c^{(b)} \sum_{x,y,z} \widetilde{p_{i,c}^{(b)}}(x,y,z) \log S_c^{(b)}(x,y,z), \tag{8}$$

where $B$ denotes the batch size, $C$ is the total number of classes, $(x, y, z)$ represents the three-dimensional voxel coordinates，and $w_c^{(b)}$ reflects the degree of class imbalance, with $w_c^{(b)} = 1 - \frac{\sum_{x,y,z}\widetilde{p_{i,c}^{(b)}}(x,y,z)}{\sum_{j=1}^{C}\sum_{x,y,z}\widetilde{p_{i,c}^{(b)}}(x,y,z)}$. During training, the random masking matrix $\mathcal{M}_{w_f}$ is re-sampled at each iteration according to the current batch's confidence $w_f$ and mask ratio, ensuring stochastic regularization across iterations.

This soft-label loss provides an additional supervision signal beyond the primary segmentation loss. Through its back propagation, the model's main encoder-decoder segmentation network learns the fused information from all modalities, explicitly promoting multi-modal fusion. Simultaneously, the back propagation optimizes the parameters within the MetaNetwork. This allows the MetaNetwork to learn how to formulate more advantageous multi-modal fusion strategies for different types of input data, generating more effective meta-parameters to guide the AMF's dynamic fusion process.

### 3.3. Consistency Regularization for Self-Supervised Learning

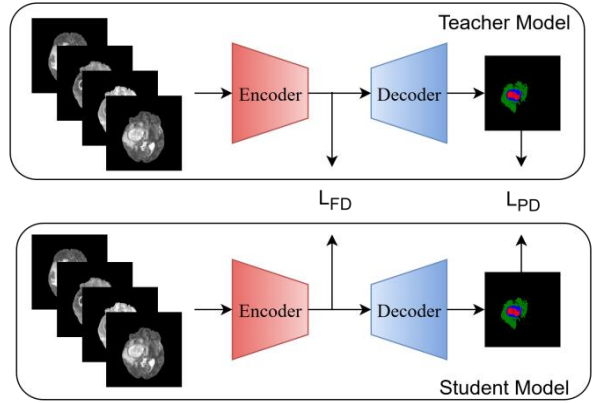

Figure 6. A simple illustration of the teacher-student paradigm in knowledge distillation. $L_{FD}$ denotes the Feature Distillation Loss, and $L_{PD}$ denotes the Prediction Distillation Loss.

**Teacher-Student Paradigm.** In traditional knowledge distillation methods [25], [26], [27], the teacher model is typically a pre-trained model with fixed parameters. However, this introduces additional training costs and computational burdens, especially for difficult tasks or those with strict accuracy requirements, as a large teacher model requires a significant number of training iterations.

Inspired by the Mean Teacher method [32] from semi-supervised learning and building on the theory of self-distillation [33], we propose a dynamically switching teacher-student-like paradigm(see Figure 6). During a single forward pass, the model, when processing a missing modality combination, acts as a "student" with less accumulated knowledge, outputting an immature prediction. Conversely, when processing the complete modality combination, it acts as a more knowledgeable "teacher," outputting a stable learning target. By leveraging the supervision signal from the complete modality input, the model can generate more accurate prediction results, even with missing modality combinations.

**Consistency Measurement and Loss Computation.** As shown in Figure 2 and Figure 4, for the current input data with a missing modality combination, we simultaneously use a complete multi-modal input as a stable learning target, where its gradients are detached. The consistency-regularized self-supervised loss is then computed from the prediction results of both inputs. Building upon [25], we align the per-pixel class probability distributions from the student to the teacher. In our implementation, the teacher and student networks share the same parameters. During training, the teacher branch is excluded from gradient computation to avoid backpropagation, thereby providing a stable learning target. This design provides a lightweight form of self-distillation without introducing additional memory or update overhead from an EMA teacher. To ensure stable training and prevent self-reinforcement of noisy predictions, the teacher outputs are generated from complete modalities under stronger augmentation, while the student learns from incomplete modalities. This asymmetry provides natural regularization and prevents representation collapse. The calculation method is as follows:

$$\mathcal{L}_{CR} = \frac{1}{DHW} \sum_{d=1}^{D} \sum_{h=1}^{H} \sum_{w=1}^{W} \mathrm{KL}\left( \phi\left(\frac{Y_t^{d,h,w}}{\tau}\right) \bigg| \phi\left(\frac{Y_s^{d,h,w}}{\tau}\right) \right), \qquad (9)$$

where $Y_s = F_\theta(X, \mathcal{M}_{\mathrm{partial}})$ and $Y_t = F_\theta(X, \mathcal{M}_{\mathrm{full}})$ represent the output results of model $F_\theta$ for the incomplete and complete modality combinations, respectively. $\mathcal{M}_{\mathrm{partial}}$ and $\mathcal{M}_{\mathrm{full}}$ are the missing and full masks applied to the input feature map. $\phi(\cdot)$ denotes the softmax activation function, KL denotes the Kullback-Leibler divergence, and $\tau$ is the temperature coefficient. By adjusting the temperature parameter t, we can convey richer inter-class relationship information, which provides more fine-grained guidance.$(d, h, w)$ represents the three-dimensional voxel coordinates.

$Y_t$ serves as an additional supervision signal, considered a gold-standard prediction for the current data point under ideal conditions. It provides a stable learning target for incomplete multi-modal combinations, allowing the model to gain valuable insights from the output of the complete multi-modal data in addition to the ground-truth labels. Consequently, when faced with missing modality data, the model can produce segmentation results that closely approximate those from a complete multi-modal input, thereby enhancing its robustness and generalization ability.

In practice, the consistency-regularized self-supervised loss ($\mathcal{L}_{CR}$) further promotes improvements in model segmentation accuracy. By minimizing the KL divergence with the teacher's prediction, the student model is compelled to learn how to utilize the available incomplete information to infer and compensate for the effects of the missing data. This implicitly promotes the model's ability to integrate multi-modal information.

Furthermore, consistent with the design philosophy of the Meta-AMF, this loss does not introduce any extra parameters or alter the original model architecture. It can be conveniently embedded into the training pipeline, acting as an effective regularization method to achieve self-reinforced learning.

### 3.4. Total Loss

As shown in Figure 2, the final total loss ($\mathcal{L}_{total}$) is a combination of the soft-label distillation loss ($\mathcal{L}_{SL}$), the consistency regularization self-supervised loss ($\mathcal{L}_{CR}$) introduced by our method, and the segmentation loss ($\mathcal{L}_{seg}$) corresponding to the primary task:

$$\mathcal{L}_{total} = \lambda_1 \mathcal{L}_{seg} + \lambda_2 \mathcal{L}_{SL} + \lambda_3 \mathcal{L}_{CR}. \tag{10}$$

Here, $\lambda_2 = 0.3 \cdot \frac{\tau^2}{10}$, $\tau$ is the temperature coefficient.

In addition to the primary segmentation loss, $\mathcal{L}_{SL}$ and $\mathcal{L}_{CR}$ serve as complementary supervision signals. These signals leverage the intrinsic knowledge within the model and the information between modalities to a greater extent. This jointly enhances the model's ability to integrate multi-modal fusion information, thereby improving its robustness and generalization when faced with missing modality combinations and ultimately boosting segmentation accuracy.

## 4. Experiments

### 4.1. Implementation Details

Our model architecture is based on the 3D U-Net design presented in [29]. It consists of an independent encoder for each modality and a single shared-parameter decoder that forms a modality-specific segmentation framework. We adopt the fusion module from [7], [8] (as shown in Figure 2) to serve as the multi-modal fusion segmentation architecture and generate the final segmentation results. Our MGML framework is then integrated into this existing method.

Following[7], BraTS2020 is randomly split into training, validation, and test sets with a ratio of

219 : 50 : 100. Consistent with [23], BraTS2023 is randomly split into training, validation, and test sets with a ratio of 7:1:2. For the performance comparison, the results of SOTA methods on BraTS2020 are directly cited from [23] and [40], as these works report evaluations under the same dataset, split, and missing-modality configurations. In contrast, for BraTS2023, we conducted new experiments under the same training and evaluation settings and in the same experimental environment. All experiments are conducted with fixed random seeds and deterministic training, ensuring reproducible Dice scores without inter-run variation.

During training, input images are randomly cropped to $80 \times 80 \times 80$ and augmented with random rotations (up to $10°$), intensity changes with a scale of (0.1, 0.1), and random flipping. We train our network for 900 epochs with a batch size of 1, where each epoch contains 150 iterations. The training was conducted on an NVIDIA RTX 3090 GPU using PyTorch version 2.7.1+cu126. Adam optimizer is employed with $\beta_1 = 0.9, \beta_2 = 0.999, \epsilon = 1 \times 10^{-8}$, and radiogram enabled. The learning rate is initialized to $2 \cdot 10^{-4}$ with weight decay $10^{-4}$, and is scheduled using the poly policy: $lr = lr_{\text{init}} \cdot \left(1 - \frac{epoch}{max\_epoch}\right)^{0.9}$. We set the random seed to 1024 to ensure reprehensibility. In addition, we empirically set $\lambda_1 = 1$, $\tau = 6$, and $\lambda_3 = 0.01$ to balance the contributions of the primary segmentation loss ($\mathcal{L}_{seg}$), the soft-label distillation loss ($\mathcal{L}_{SL}$), and the consistency regularization self-supervised loss ($\mathcal{L}_{CR}$), respectively. For parameter initialization, we adopt Kaiming normal initialization for all convolutional layers. The average training time per epoch is reported in Table 7 for both baseline models (RFNet and mmFormer) with and without the MGML module.

### 4.2. Dataset

We evaluate our method using the BraTS2020 dataset from the Multimodal Brain Tumor Segmentation Challenge [1]. This dataset contains 369 subjects, each with four distinct MRI modalities: FLAIR, T1-weighted with contrast enhancement (T1ce), T1-weighted (T1), and T2-weighted (T2). Consistent with the splitting scheme used in RFNet [7], we divide the dataset into 219 subjects for training, 50 for validation, and 100 for testing.

In BraTS2020, all images were skull-stripped, resampled to an isotropic 1 mm³ resolution, and co-registered to the same anatomical template, ensuring spatial alignment across modalities. Expert annotations were provided for three clinically relevant tumor subregions: the enhancing tumor (ET), the tumor core (TC, including ET and necrotic/non-enhancing regions), and the whole tumor (WT, including TC and peritumoral edema). The dataset is widely regarded as a benchmark for multimodal brain tumor segmentation due to its multi-institutional origin, heterogeneous acquisition protocols, and carefully curated ground truth labels, which collectively pose significant challenges for robust and generalizable model development.

### 4.3. Evaluation Metric

The dice coefficient [34] is used to measure the segmentation performance of our proposed method. It is defined as:

$$Dice_{\bar{k}}(\hat{y}, y) = \frac{2 \cdot |\widehat{y_{\bar{k}}} \cap y_{\bar{k}}|_1}{|\widehat{y_{\bar{k}}}|_1 + |y_{\bar{k}}|_1}, \tag{11}$$

where k denotes the tumor classes, including WT, TC, and ET. The dice score, $Dice_{\bar{k}}$, represents the segmentation accuracy for tumor class k. A larger dice score indicates that the predictions are more like the ground truth, thus signifying better segmentation performance.

**4.4. Ablation Study**

Table 1. Our proposed method (MGML)'s ablation experiment on the baseline.

| M | FLAIR | ✓ | ✗ | ✗ | ✗ | ✓ | ✓ | ✓ | ✗ | ✗ | ✗ | ✓ | ✓ | ✓ | ✗ | ✓ | AVG |
|---|---|---|---|---|---|---|---|---|---|---|---|---|---|---|---|---|---|
| | T1ce | ✗ | ✓ | ✗ | ✗ | ✓ | ✗ | ✗ | ✓ | ✓ | ✗ | ✓ | ✓ | ✗ | ✓ | ✓ | |
| | T1 | ✗ | ✗ | ✓ | ✗ | ✗ | ✓ | ✗ | ✓ | ✗ | ✓ | ✓ | ✗ | ✓ | ✓ | ✓ | |
| | T2 | ✗ | ✗ | ✗ | ✓ | ✗ | ✗ | ✓ | ✗ | ✓ | ✓ | ✗ | ✓ | ✓ | ✓ | ✓ | |
| WT | Baseline | 85.29 | 69.35 | 65.56 | 83.12 | 88.08 | 88.00 | 88.71 | 75.92 | 85.57 | 85.55 | 88.86 | 89.07 | 89.33 | 86.47 | 89.68 | 83.90 |
| | Baseline + $\mathcal{L}_{CR}$ | 85.07 | 72.43 | 71.12 | 84.20 | 88.07 | 87.94 | 88.55 | 77.35 | 86.30 | 86.29 | 88.74 | 89.21 | 89.07 | 86.80 | 89.44 | 84.70 |
| | Baseline + Meta-AMF | 85.81 | 73.03 | 73.50 | 84.03 | 88.76 | 88.78 | 89.08 | 77.97 | 86.38 | 86.59 | 89.41 | 89.77 | 89.77 | 87.21 | 90.21 | 85.35 |
| | Baseline + MGML | **86.84** | **75.07** | **75.92** | **85.66** | **89.27** | **89.31** | **89.43** | **80.29** | **87.28** | **87.42** | **89.77** | **90.09** | **90.08** | **87.90** | **90.42** | **86.31** |
| TC | Baseline | 63.57 | 73.03 | 55.96 | 66.89 | 80.56 | 68.25 | 70.30 | 79.48 | 81.13 | 69.36 | 82.32 | 82.02 | 71.64 | 82.28 | 82.78 | 73.97 |
| | Baseline + $\mathcal{L}_{CR}$ | 64.63 | 79.82 | 58.18 | 68.31 | 83.14 | 69.58 | 71.01 | 82.03 | 82.90 | 70.34 | 83.72 | 83.63 | 72.42 | 83.28 | 83.90 | 75.79 |
| | Baseline + Meta-AMF | 66.19 | **80.42** | 60.57 | 70.04 | 83.19 | 70.42 | **72.69** | 82.35 | 83.10 | 72.19 | 83.49 | **84.09** | 73.70 | 83.53 | 84.09 | 76.67 |
| | Baseline + MGML | **66.79** | 80.03 | **62.38** | **70.57** | **83.65** | **72.07** | 72.68 | **82.84** | **83.16** | **72.43** | **83.79** | 84.05 | **74.06** | **83.67** | **84.10** | **77.08** |
| ET | Baseline | 36.29 | 66.24 | 29.66 | 41.93 | 71.11 | 39.83 | 42.76 | 70.47 | 72.07 | 43.82 | 74.82 | 73.61 | 44.70 | 74.90 | 73.98 | 57.07 |
| | Baseline + $\mathcal{L}_{CR}$ | 35.04 | 72.19 | 29.81 | 43.08 | 73.20 | 39.20 | 44.22 | 73.30 | 75.44 | 44.02 | 73.61 | 74.61 | 45.06 | 75.91 | 75.06 | 58.24 |
| | Baseline + Meta-AMF | 36.39 | **74.39** | 33.49 | 43.57 | 75.17 | 41.08 | 45.55 | 76.75 | 75.46 | 45.53 | 75.81 | 75.69 | 46.24 | 76.03 | 76.26 | 59.82 |
| | Baseline + MGML | **36.61** | 73.90 | **34.22** | **45.41** | **75.50** | **41.85** | **46.18** | **77.82** | **75.74** | **46.40** | **76.29** | **75.86** | **47.50** | **76.13** | **76.35** | **60.38** |

As shown in Table 1, the introduction of the Meta-AMF module led to improvements of 1.45%, 2.70%, 2.75%, and 2.30% in the dice scores for WT, TC, and ET, as well as the overall average (AVG), respectively. With the further integration of $\mathcal{L}_{CR}$, we observed additional performance gains of 0.96%, 0.41%, 0.56%, and 0.64% for the same metrics. At the same time, it can be observed that under our MGML framework, we achieve a comprehensive improvement in WT segmentation across all modality combinations when built upon the baseline. For TC and ET, although our method may not achieve the best performance in a few individual modality combinations, it still delivers results that are very close to the optimal performance, and it achieves an overall improvement in the AVG score across all combinations. This validates the effectiveness of our framework in enhancing the baseline's performance.

Table 2. Ablation study results for fixed-parameter soft label distillation (FPSLD) and Meta-AMF

| Methods | WT | TC | ET | AVG |
|---|---|---|---|---|
| Baseline | 83.90 | 73.97 | 57.07 | 71.64 |
| Baseline + FPSLD | 85.15 | 76.28 | 59.43 | 73.62 |
| Baseline + Meta-AMF | **85.35** | **76.67** | **59.82** | **73.95** |

As shown in Table 2, we tested Fixed-Parameter Soft Label Distillation (FPSLD) on the baseline by fixing the meta-parameters $\alpha$ and $\beta$ to 100 and the fusion balance weight to 0.5. This approach aggressively obtains soft labels from the per-modality predictions of the shared decoder, and the

MetaNetwork does not adaptively update its parameters to adjust the fusion strategy during training. The results show that our proposed Meta-AMF demonstrates greater superiority over FPSLD, achieving further improvements of 0.2%, 0.29%, 0.39%, and 0.33% on WT, TC, and ET, and the overall average (AVG), respectively. This highlights the necessity of dynamically adjusting the fusion balance.

### 4.5. Comparisons with the State-of-the-art

Table 3. Performance comparison (DSC%) with SOTA methods on BraTS2020. Available and missing modalities are denoted by ✓ and ✗, respectively.

| M | Method | FLAIR ✓ T1ce ✗ T1 ✗ T2 ✗ | ✗ ✓ ✗ ✗ | ✗ ✗ ✓ ✗ | ✗ ✗ ✗ ✓ | ✓ ✓ ✗ ✗ | ✓ ✗ ✓ ✗ | ✓ ✗ ✗ ✓ | ✗ ✓ ✓ ✗ | ✗ ✓ ✗ ✓ | ✗ ✗ ✓ ✓ | ✓ ✗ ✓ ✓ | ✓ ✓ ✗ ✓ | ✓ ✓ ✓ ✗ | ✗ ✓ ✓ ✓ | ✓ ✓ ✓ ✓ | AVG |
|---|---|---|---|---|---|---|---|---|---|---|---|---|---|---|---|---|---|
| WT | HeMiS[35] (MICCAI,2016) | 71.60 | 67.71 | 68.96 | 68.19 | 69.17 | 68.67 | 69.83 | 69.01 | 69.78 | 69.40 | 70.21 | 71.28 | 70.73 | 71.58 | 72.06 | 69.88 |
| | U-HVED[36] (MICCAI,2019) | 69.85 | 46.82 | 46.77 | 54.03 | 61.45 | 58.25 | 64.50 | 62.91 | 65.76 | 64.29 | 66.99 | 69.70 | 68.38 | 70.35 | 71.41 | 62.76 |
| | RFNet[7] (ICCV,2021) | 86.42 | 77.34 | 76.46 | 86.21 | 89.55 | 89.30 | 89.35 | 81.00 | 87.45 | 87.95 | 90.39 | 90.20 | 90.42 | 88.59 | 90.77 | 86.76 |
| | UNet-MFI[37] (MICCAI,2022) | 82.27 | 73.18 | 72.10 | 82.45 | 83.64 | 84.34 | 84.85 | 77.30 | 83.44 | 83.52 | 85.45 | 85.70 | 85.85 | 84.20 | 84.93 | 82.21 |
| | mmFormer[8] (MICCAI,2022) | 82.40 | 74.25 | 74.37 | 83.07 | 84.54 | 84.61 | 85.82 | 77.98 | 84.05 | 84.00 | 85.34 | 86.11 | 86.22 | 84.64 | 86.38 | 82.92 |
| | MMANet[38] (CVPR,2023) | 83.21 | 64.46 | 64.35 | 82.88 | 86.74 | 87.64 | 87.47 | 70.98 | 85.64 | 85.33 | 87.92 | 89.37 | 87.97 | 86.38 | 89.25 | 82.64 |
| | PASSION[39] (ACM MM,2024) | 70.51 | 70.69 | 84.22 | 83.58 | 76.30 | 87.71 | 85.62 | **86.90** | 86.26 | 87.83 | 88.65 | 88.00 | 88.82 | 89.36 | 89.74 | 84.28 |
| | DMAF-Net[40] (arXiv, 2025) | 72.92 | 70.64 | **84.94** | 83.64 | 77.57 | 87.60 | 85.86 | **86.90** | 86.67 | **88.25** | 88.30 | 88.27 | 89.53 | **89.91** | 89.60 | 84.71 |
| | IM-Fuse [23] (MICCAI,2025) | 87.55 | 76.90 | 77.52 | 86.29 | 89.57 | 89.81 | 89.90 | 81.28 | 88.01 | 88.03 | 90.44 | 90.44 | 90.73 | 88.79 | 91.02 | 87.08 |
| | mmFormer+MGML | 86.84 | 75.07 | 75.92 | 85.66 | 89.27 | 89.31 | 89.43 | 80.29 | 87.28 | 87.42 | 89.77 | 90.09 | 90.08 | 87.90 | 90.42 | 86.31 |
| | RFNet+MGML | 87.79 | 77.86 | 77.89 | 86.57 | 89.64 | 89.75 | 89.90 | 81.93 | 88.20 | 88.04 | 90.53 | **90.79** | 90.35 | 89.00 | 91.03 | 87.28 |
| | IM-Fuse+MGML | **88.22** | **78.08** | 78.74 | **86.80** | **89.95** | **89.93** | **90.11** | 82.26 | **88.46** | 88.14 | **90.73** | 90.61 | **90.98** | 89.12 | **91.21** | **87.55** |
| TC | HeMiS[35] (MICCAI,2016) | 53.43 | 51.41 | 51.56 | 51.11 | 51.70 | 51.08 | 51.85 | 51.88 | 52.35 | 51.51 | 52.95 | 53.76 | 52.97 | 54.38 | 55.03 | 52.46 |
| | U-HVED[36] (MICCAI,2019) | 34.62 | 35.51 | 27.30 | 37.67 | 42.15 | 38.26 | 43.41 | 44.93 | 47.53 | 44.97 | 49.13 | 51.30 | 49.40 | 52.72 | 54.17 | 43.53 |
| | RFNet[7] (ICCV,2021) | 65.04 | 82.37 | 64.31 | 68.47 | 84.69 | 71.45 | 72.62 | 83.15 | 84.06 | 72.11 | 84.71 | 84.70 | 74.28 | 84.11 | 84.74 | 77.39 |
| | UNet-MFI[37] (MICCAI,2022) | 63.94 | 77.63 | 59.38 | 68.05 | 79.92 | 68.23 | 70.72 | 77.61 | 80.09 | 70.21 | 80.03 | 80.94 | 71.40 | 80.75 | 81.28 | 74.01 |
| | mmFormer[8] (MICCAI,2022) | 66.19 | 77.96 | 61.17 | 69.18 | 80.36 | 69.58 | 71.55 | 79.93 | 80.79 | 70.90 | 80.18 | 81.31 | 72.02 | 81.12 | 81.22 | 74.90 |
| | MMANet[38] (CVPR,2023) | 65.79 | 74.05 | 57.47 | 70.21 | **85.21** | 72.73 | 72.65 | 80.99 | 85.24 | 71.88 | **85.87** | 85.68 | 74.48 | **86.39** | **86.63** | 76.99 |
| | PASSION[39] (ACM MM,2024) | 55.47 | 73.11 | 59.50 | 61.76 | 78.21 | 66.17 | 65.98 | 76.33 | 78.74 | 67.06 | 79.51 | 80.54 | 68.62 | 79.60 | 79.93 | 71.37 |
| | DMAF-Net[40] (arXiv, 2025) | 56.02 | 74.31 | 60.55 | 63.18 | 78.63 | 66.31 | 66.42 | 76.85 | 79.98 | 67.34 | 79.44 | 81.06 | 68.84 | 80.30 | 80.47 | 71.98 |
| | IM-Fuse [23] (MICCAI,2025) | 70.18 | 83.09 | 66.46 | 71.22 | 85.15 | 73.90 | 73.81 | 84.64 | 85.46 | 73.88 | 85.50 | 85.21 | 75.50 | 85.74 | 85.51 | 79.01 |
| | mmFormer+MGML | 66.79 | 80.03 | 62.38 | 70.57 | 83.65 | 72.07 | 72.68 | 82.84 | 83.16 | 72.43 | 83.79 | 84.05 | 74.06 | 83.67 | 84.10 | 77.08 |
| | RFNet+MGML | 71.04 | **83.62** | 66.52 | 71.52 | 85.01 | 74.13 | 74.08 | **85.16** | 85.71 | **74.07** | 85.55 | 85.28 | 75.65 | 86.00 | 85.63 | 79.26 |
| | IM-Fuse+MGML | **71.20** | 83.44 | **67.29** | **71.89** | 85.17 | **74.70** | **74.66** | 84.84 | **85.78** | 73.90 | 85.41 | 84.34 | **75.82** | 85.52 | 85.46 | **79.36** |
| ET | HeMiS[35] (MICCAI,2016) | **43.77** | 42.41 | 41.59 | 41.45 | 41.83 | 40.29 | 41.19 | 42.08 | 42.39 | 41.00 | 43.67 | 44.16 | 42.95 | 45.27 | 46.33 | 42.69 |
| | U-HVED[36] (MICCAI,2019) | 12.88 | 24.94 | 7.27 | 24.26 | 30.02 | 21.95 | 29.40 | 33.64 | 36.18 | 32.12 | 39.39 | 40.91 | 38.09 | 43.18 | 45.33 | 30.64 |
| | RFNet[7] (ICCV,2021) | 40.47 | **74.27** | 37.51 | 43.59 | 76.45 | 43.81 | 46.99 | 75.22 | 73.94 | 46.37 | **77.01** | 76.38 | 48.95 | 76.38 | 76.64 | 60.93 |
| | UNet-MFI[37] (MICCAI,2022) | 39.70 | 69.42 | 29.38 | 46.00 | 70.13 | 40.06 | 48.69 | 69.25 | 72.32 | 45.71 | 71.28 | 70.88 | 46.55 | 72.00 | 71.41 | 57.52 |
| | mmFormer[8] (MICCAI,2022) | 40.47 | 68.91 | 33.97 | 45.61 | 69.81 | 43.63 | 48.09 | 71.10 | 70.72 | 45.92 | 70.08 | 71.60 | 48.38 | 70.65 | 71.36 | 58.02 |
| | MMANet[38] (CVPR,2023) | 36.40 | 66.07 | 28.22 | 41.91 | 70.43 | 41.50 | 44.23 | 68.87 | 71.86 | 43.58 | 71.94 | 71.98 | 44.60 | 71.81 | 72.43 | 56.39 |
| | PASSION[39] （ACM MM,2024） | 28.47 | 66.71 | 31.25 | 37.12 | 69.58 | 38.09 | 40.66 | 68.27 | 67.86 | 40.85 | 69.27 | 69.90 | 42.94 | 68.97 | 69.77 | 53.92 |
| | DMAF-Net[40] （arXiv, 2025） | 30.37 | 66.97 | 31.64 | 36.86 | 70.37 | 38.20 | 41.66 | 68.62 | 68.02 | 40.33 | 69.50 | 70.11 | 43.35 | 69.06 | 69.45 | 54.30 |
| | IM-Fuse [23] （MICCAI,2025） | 41.21 | 73.11 | 36.25 | 47.37 | 74.81 | 45.41 | 49.09 | 76.48 | 76.06 | 49.24 | 76.68 | 76.29 | 50.03 | 77.00 | 76.63 | 61.70 |
| | mmFormer+MGML | 36.61 | 73.90 | 34.22 | 45.41 | 75.50 | 41.85 | 46.18 | 77.82 | 75.74 | 46.40 | 76.29 | 75.86 | 47.50 | 76.13 | 76.35 | 60.38 |
| | RFNet+MGML | 40.78 | 73.63 | 36.66 | 49.03 | 73.88 | 45.29 | 49.05 | 76.94 | 76.95 | 48.95 | 75.59 | 76.14 | **51.79** | 78.37 | **78.69** | 62.11 |
| | IM-Fuse+MGML | 41.36 | 73.90 | **37.77** | **49.23** | 76.11 | **46.76** | **49.17** | **77.84** | **77.25** | **49.01** | 75.56 | **77.33** | 51.68 | **79.48** | 77.72 | **62.67** |

As shown in Table 3, to validate the effectiveness of our MGML framework, we compared its performance against several state-of-the-art (SOTA) methods from recent years, including HeMis [35],U-HVED [36], RFNet[7], UNet-MFI [37], mmFormer [8], MMANet [38], PASSION [39] DMAFNet [40] and IM-Fuse[23]. In a comparison across fifteen missing modality combinations for WT, TC, and ET, our method achieved superior performance under most conditions.

Specifically, HeMIS [35] introduced one of the earliest frameworks for handling missing modalities by learning modality-invariant representations through a hetero-modal segmentation network. U-HVED [36] extended this idea with a variational encoder-decoder to jointly complete missing modalities and perform segmentation. UNet-MFI [37] proposed a modality-adaptive feature interaction mechanism to dynamically model feature dependencies across available modalities. RFNet [7] proposed a hierarchical pyramid structure that, through its region-aware capability for brain tumors. mmFormer [8] leveraged transformer-based cross-modal attention to enhance incomplete multimodal learning. MMANet [38] incorporated margin-aware distillation and modality-aware regularization to explicitly handle missing modality scenarios. PASSION [39] designed a strategy to cope with imbalanced missing rates by selectively enhancing underrepresented modality combinations. DMAF-Net [40] introduced a modality rebalancing framework to dynamically adjust feature contributions across incomplete modalities. Finally, IM-Fuse[23] recently introduced a Mamba-based latent-space fusion strategy, where the Interleaved Mamba Fusion Block (I-MFB) enables selective propagation of multimodal features.

By integrating MGML into diverse baseline architectures, including CNN-based (RFNet) and hybrid CNN-transformer or CNN–state-space architectures (mmFormer and IM-Fuse), we observe consistent and substantial performance improvements across nearly all missing-modality combinations. Specifically, MGML enhances RFNet by 0.52%, 1.87%, and 1.18% in WT, TC, and ET, respectively; boosts mmFormer by 3.39%, 2.18%, and 2.36%; and improves IM-Fuse by 0.47%, 0.35%, and 0.97% in the same metrics. These consistent gains across architectures with different design paradigms clearly demonstrate the versatility, robustness, and plug-and-play nature of the proposed MGML framework.

Table 4. Performance comparison (DSC%) with SOTA methods on BraTS2023. Available and missing modalities are denoted by

✓ and ✗, respectively.

| M | | FLAIR | ✓ | ✗ | ✗ | ✗ | ✓ | ✓ | ✓ | ✗ | ✗ | ✗ | ✓ | ✓ | ✓ | ✗ | ✓ | |
|---|---|---|---|---|---|---|---|---|---|---|---|---|---|---|---|---|---|---|
| | | T1ce | ✗ | ✓ | ✗ | ✗ | ✓ | ✗ | ✗ | ✓ | ✓ | ✗ | ✓ | ✓ | ✗ | ✓ | ✓ | |
| | | T1 | ✗ | ✗ | ✓ | ✗ | ✗ | ✓ | ✗ | ✓ | ✗ | ✓ | ✓ | ✗ | ✓ | ✓ | ✓ | AVG |
| | | T2 | ✗ | ✗ | ✗ | ✓ | ✗ | ✗ | ✓ | ✗ | ✓ | ✓ | ✗ | ✓ | ✓ | ✓ | ✓ | |
| WT | U-HVED[36] (MICCAI,2019) | | 83.12 | 77.28 | 73.52 | 80.18 | 81.64 | 82.98 | 79.63 | 70.35 | 82.68 | 78.45 | 82.12 | 83.65 | 83.16 | 78.23 | 84.45 | 80.10 |
| | mmFormer[8] (MICCAI,2022) | | 85.87 | 78.21 | 78.64 | 85.11 | 88.05 | 88.33 | 88.57 | 81.79 | 87.12 | 86.93 | 90.18 | 90.34 | 90.02 | 88.79 | 91.05 | 86.60 |
| | RFNet[7] (ICCV,2021) | | 87.92 | 81.04 | 80.91 | 87.35 | 89.68 | 89.95 | 90.08 | 83.19 | 88.34 | 88.11 | 91.21 | 91.44 | 91.16 | 89.84 | 91.95 | 88.14 |
| | IM-Fuse [23] (MICCAI,2025) | | 89.48 | 82.22 | 82.01 | 88.94 | 91.33 | 91.74 | 92.15 | 84.76 | 89.89 | 89.47 | 92.73 | 93.11 | 92.55 | 91.46 | 93.62 | 89.69 |
| | mmFormer+MGML | | 87.95 | 79.95 | 79.92 | 86.92 | 89.87 | 90.41 | 90.69 | 83.65 | 88.91 | 88.62 | 91.72 | 91.95 | 91.73 | 90.33 | 92.56 | 88.35 |
| | RFNet+MGML | | 88.67 | 81.60 | 81.32 | 88.04 | 90.49 | 90.91 | 91.12 | 84.22 | 89.32 | 88.97 | 92.14 | 92.36 | 92.01 | 90.85 | 93.01 | 89.00 |
| | IM-Fuse+MGML | | **90.95** | **82.83** | **82.60** | **89.42** | **91.92** | **92.48** | **92.71** | **85.73** | **90.81** | **90.49** | **93.43** | **93.81** | **93.34** | **92.12** | **94.45** | **90.47** |
| TC | U-HVED[36] (MICCAI,2019) | | 62.78 | 70.56 | 60.18 | 65.78 | 74.56 | 65.72 | 64.12 | 78.58 | 76.73 | 64.19 | 79.96 | 77.34 | 64.84 | 76.32 | 75.75 | 70.49 |
| | mmFormer[8] (MICCAI,2022) | | 70.95 | 82.01 | 65.88 | 70.91 | 83.98 | 73.86 | 73.75 | 84.05 | 84.52 | 73.58 | 85.08 | 85.11 | 74.96 | 85.97 | 86.48 | 78.74 |
| | RFNet[7] (ICCV,2021) | | 72.86 | 83.56 | 67.17 | 72.21 | 85.22 | 75.12 | 74.92 | 85.18 | 85.64 | 74.53 | 86.22 | 86.05 | 76.09 | 86.98 | 87.51 | 79.95 |
| | IM-Fuse [23] (MICCAI,2025) | | 74.91 | 84.88 | 69.31 | 74.01 | 87.22 | 77.03 | 76.65 | 86.64 | 87.93 | 76.02 | 87.77 | 87.31 | 78.64 | 88.22 | 89.54 | 81.74 |
| | mmFormer+MGML | | 72.98 | 84.12 | 67.91 | 72.82 | 85.84 | 75.92 | 75.82 | 85.86 | 86.23 | 75.11 | 86.87 | 86.71 | 76.71 | 87.62 | 88.11 | 80.58 |
| | RFNet+MGML | | 74.02 | 85.12 | 68.92 | 73.89 | 86.71 | 76.85 | 76.51 | 86.51 | 86.95 | 75.89 | 87.45 | 87.28 | 77.38 | 88.27 | 88.62 | 81.36 |
| | IM-Fuse+MGML | | **75.83** | **86.73** | **70.22** | **75.54** | **88.43** | **78.61** | **78.25** | **88.09** | **88.63** | **77.65** | **89.14** | **88.95** | **79.33** | **89.73** | **90.02** | **83.01** |
| ET | U-HVED[36] (MICCAI,2019) | | 39.46 | 72.45 | 31.67 | 45.12 | 72.45 | 42.16 | 48.64 | 75.35 | 74.82 | 73.69 | 71.86 | 73.12 | 50.04 | 76.14 | 74.24 | 61.41 |
| | mmFormer[8] (MICCAI,2022) | | 45.10 | 77.18 | 41.10 | 53.05 | 78.20 | 49.12 | 53.08 | 81.12 | 81.84 | 51.61 | 79.15 | 80.21 | 55.93 | 81.54 | 82.09 | 66.02 |
| | RFNet[7] (ICCV,2021) | | 48.05 | 79.90 | 42.87 | 54.81 | 79.85 | 50.78 | 54.17 | 82.85 | 82.97 | 52.72 | 80.52 | 81.61 | 57.08 | 82.92 | 83.48 | 67.64 |
| | IM-Fuse [23] (MICCAI,2025) | | 49.67 | 81.65 | 44.26 | 56.93 | 81.45 | 52.39 | 55.88 | 84.57 | 84.91 | 54.12 | 82.66 | 83.81 | 58.83 | 84.64 | 85.41 | 69.41 |
| | mmFormer+MGML | | 49.08 | 80.70 | 43.32 | 55.42 | 80.52 | 51.48 | 54.91 | 83.41 | 83.55 | 53.21 | 81.18 | 82.21 | 57.64 | 83.50 | 84.16 | 68.29 |
| | RFNet+MGML | | 50.12 | 81.05 | 43.89 | 56.05 | 81.05 | 51.97 | 55.32 | 83.99 | 84.01 | 53.78 | 81.64 | 82.76 | 58.21 | 83.99 | 84.61 | 68.83 |
| | IM-Fuse+MGML | | **51.46** | **82.59** | **45.02** | **57.53** | **82.41** | **53.36** | **56.83** | **85.54** | **85.73** | **55.27** | **83.18** | **84.39** | **59.53** | **85.46** | **86.21** | **70.30** |

To further verify the generalizability and robustness of the proposed MGML framework, we conducted additional experiments on the BraTS2023 dataset, which contains more diverse and standardized data compared with BraTS2020. We selected several representative baselines with different architectural characteristics, including RFNet, mmFormer, and IM-Fuse, and integrated MGML into each of them to assess its plug-and-play adaptability across diverse architectures. The results, summarized in Table 4, show that all MGML-enhanced models (e.g., RFNet+MGML, mmFormer+MGML, and IM-Fuse+MGML) consistently outperform their corresponding baselines across different missing-modality combinations. These findings further demonstrate the adaptability and robustness of MGML to newer data distributions and acquisition variations in the BraTS2023 dataset.

**4.6. Hyperparameters Sensitivity Analysis**

Table 5. Sensitivity analysis of the temperature coefficient $\tau$

| $\tau$ | dice (%) | | | |
|---|---|---|---|---|
| | WT | TC | ET | AVG |
| 2 | 87.12 | 79.23 | 62.03 | 76.16 |
| 4 | 87.18 | 79.16 | 62.08 | 76.14 |
| 6 | **87.28** | **79.26** | 62.11 | **76.21** |
| 8 | 87.20 | 79.25 | **62.13** | 76.19 |
| 10 | 87.15 | 79.12 | 62.05 | 76.15 |

In Section 3.4, we discussed the form of our final loss, where the weight $\lambda_2$ of the soft-label loss is highly correlated with the temperature coefficient $\tau$. Therefore, in Table 5, we analyze the impact of $\tau$ on the final segmentation results, which serves as a sensitivity analysis. It's clear that for a range of values from 2 to 10, the final dice score is not significantly affected, fluctuating by only about 0.1%. This indicates that Meta-AMF plays an adaptive regulatory role under different temperature coefficients. The best results are obtained when the temperature coefficient $\tau$ is set to 6.

Table 6. Sensitivity analysis of $\lambda_3$

| $\lambda_3$ | dice (%) | | | |
|---|---|---|---|---|
| | WT | TC | ET | AVG |
| 0.001 | 87.10 | 79.10 | 62.00 | 76.14 |
| 0.01 | **87.28** | **79.26** | **62.11** | **76.21** |
| 0.1 | 87.15 | 79.20 | 62.05 | 76.19 |

To investigate the sensitivity of the hyperparameter $\lambda_3$, we conducted experiments with three values: 0.001, 0.01, and 0.1. As shown in Table 6, setting $\lambda_3$ yields the best overall performance with Dice scores of 87.28%, 79.26%, and 62.11% for WT, TC, and ET, respectively. When $\lambda_3$ is slightly smaller (0.001) or larger (0.1), the performance decreases marginally, indicating that the model is relatively robust to variations in $\lambda_3$. This analysis demonstrates that our MGML framework maintains stable segmentation performance across a reasonable range of the regularization weight.

## 4.7. Meta-Parameter Effectiveness Analysis

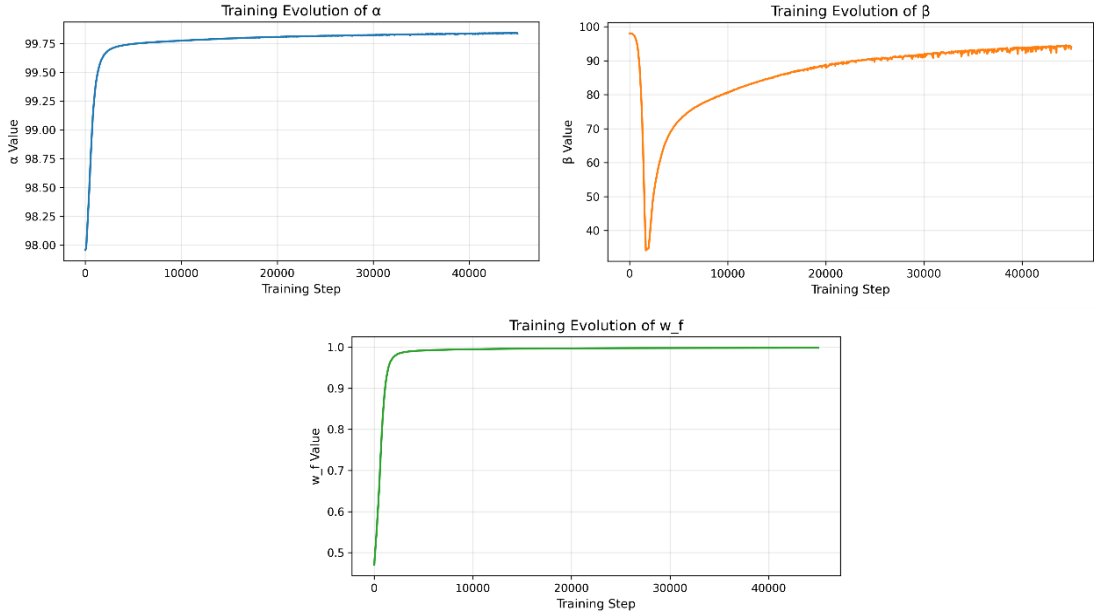

Figure 7. Evolution of the meta-parameters ($w_f$, β and α) during training.

To further illustrate the interpretability of the proposed meta-parameterized fusion mechanism, we analyze the variation of the learned meta-parameters ($w_f$, β and α) during training. These parameters are dynamically generated by the MetaNetwork to adaptively balance confident and conservative fusion behaviors under different modality conditions.

As shown in Figure 7, both α and $w_f$ exhibit stable upward trends and quickly converge to high values, indicating that the MetaNetwork learns consistent fusion preferences once reliable modality relationships are established. In contrast, β shows a clear "decrease–increase" pattern, suggesting that the model initially adopts a conservative fusion strategy and gradually shifts toward more confident modality selection as the feature representations become more discriminative. These results provide an empirical view of how the meta-parameters evolve during training, supporting the effectiveness of the MetaNetwork in dynamically adjusting fusion strategies across different modality combinations.

# 5. Discussion

## 5.1. Visualization Results of Our Method

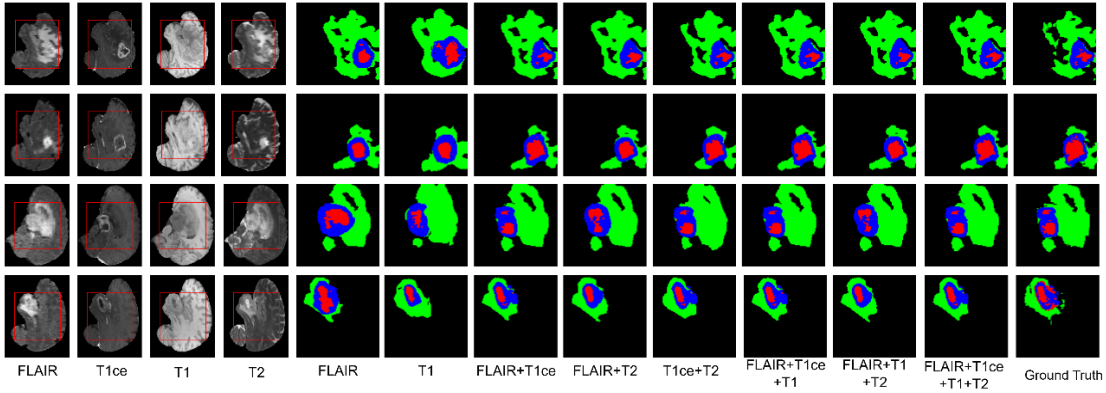

Figure 8. Predicted segmentation maps from our approach on BraTS2020. The four image modalities (T1, T1c, T2, FLAIR) are shown on the left. The segmentation maps on the right demonstrate the model's predictions under various modality-missing scenarios, compared against the ground truth.

As illustrated in Figure 8, we present representative slices from four subjects in the BraTS2020 dataset, where each row corresponds to one subject under various modality-missing scenarios. This visualization demonstrates the robustness and accuracy of our method across different subjects and incomplete modality conditions. The WT, TC, and ET are annotated in green, blue, and red, respectively. To provide a clearer view of the results, the MRI slice has been center-cropped to focus on the core segmentation region. The figure shows that even with a single-modality input, our method yields relatively accurate predictions, which highlights its strong robustness to missing modalities. With a complete multi-modal input, our approach fully leverages the fused information to produce even more precise segmentation results. Our method accurately predicts the WT and TC regions under various modality combinations. For the more challenging ET region, our method demonstrates superior performance when all modalities are present, achieving notably accurate predictions for ET areas within the TC region.

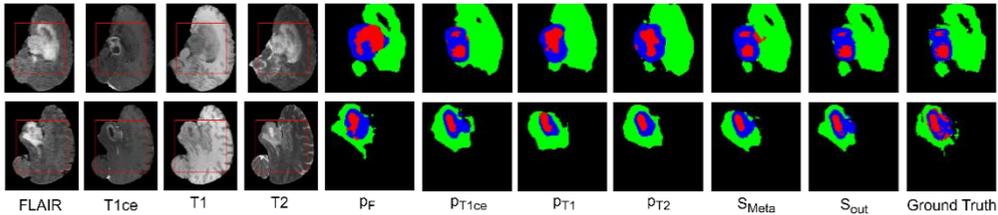

Figure 9. Visual Comparison of Segmentation Results on the BraTS2020 Dataset with All Modalities Present, along with the Predictions from the Independent Modality Encoder-Decoder and the Soft Labels Generated by the Meta-AMF.

To demonstrate the effectiveness of our MGML in using information from the independent modality encoder-decoder to generate adaptive soft labels, we selected one slice from two different subjects in the dataset. We then visualized the predictions from the independent modality encoder-decoder ($P_F$, $P_{T1ce}$,

$P_{T1}$, and $P_{T2}$), the Meta-AMF Output ($S_{meta}$), the Fusion Prediction ($S_{out}$), and the Ground Truth.

As shown in Figure 9, all predictions are based on a complete multi-modal input. The predictions from each individual independent modality encoder-decoder have their own strengths. For instance, FLAIR accurately predicts voids within the large whole tumor (WT) region, while T1ce precisely segments the enhancing tumor (ET) within the tumor core (TC) region into two parts. However, none of them achieve a comprehensive advantage on their own. Our MGML effectively uses the outputs of each independent encoder-decoder for adaptive multi-modal fusion weighting. By adopting a more effective fusion strategy for different modality inputs, it generates higher-quality soft labels, providing an additional supervision signal for the training process. As a result, the final fusion prediction achieves a segmentation accuracy that is very close to the Ground Truth. This is also supported by the results in Figure 8, which show that our method achieves relatively accurate segmentation even with incomplete modality inputs.

## 5.2. Comparisons with Other Methods

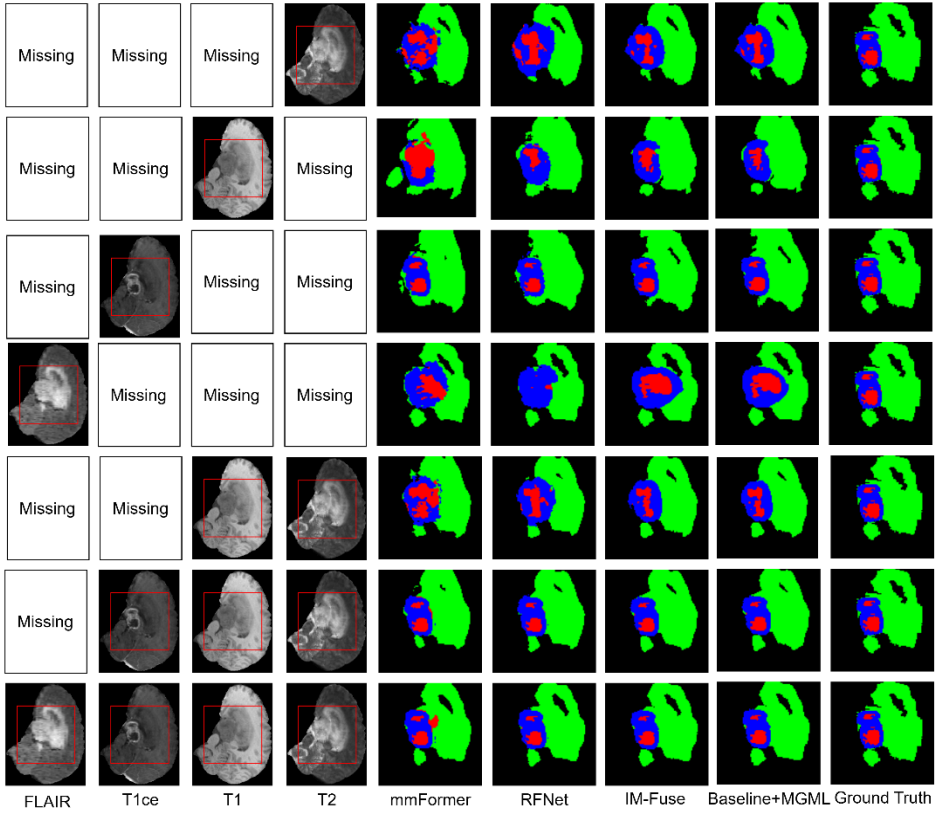

Figure 10. Visual comparison of segmentation results from different methods on the BraTS2020 dataset, across various available modalities.

As summarized in Table 3, we compare our method against a wide range of state-of-the-art (SOTA) approaches for incomplete multi-modal brain tumor segmentation. These methods can be broadly categorized into three groups: Modality completion or synthesis-based approaches (e.g., HeMIS [35], HVED [36], PASSION [39], DMAF-Net [40]), which attempt to impute or rebalance missing modalities. While effective in alleviating data incompleteness, their reliance on modality restoration introduces risks of error propagation and struggles in handling highly imbalanced missing patterns. Knowledge-distillation-based methods (e.g., Wang et al. [11], Mmanet [38]), which transfer knowledge between modalities or across teacher–student models. However, these methods often require external teacher networks or rely on static modality weights, limiting their flexibility and generalization. Fusion-driven methods (e.g., RFNet [7], mmFormer [8], and IM-Fuse[23]), which directly focus on feature or representation fusion. Although they alleviate the burden of explicit modality completion, existing designs usually adopt fixed or heuristic fusion strategies, making them less adaptive to the heterogeneous and uncertain interactions across modalities.

To address these limitations, we propose the meta-guided multi-modal learning (MGML) framework, which introduces two complementary modules: The Meta-AMF module dynamically generates meta-parameters tailored for each input, enabling a more flexible balance between optimistic and conservative fusion, thus providing reliable soft supervisory signals beyond fixed fusion rules. The consistency regularization module adopts a teacher–student paradigm using the model's own complete and incomplete states, offering auxiliary self-supervision without the need for an external teacher model. This design not only enhances robustness but also implicitly improves multimodal integration.

Building upon baseline networks, our MGML consistently achieves superior segmentation performance across almost all missing-modality settings, as shown in Table 3. Notably, the gains are particularly pronounced for the enhancing tumor (ET) region, which is notoriously challenging due to its unstable appearance and clinical ambiguity.

To further highlight the practical effectiveness of our framework, we conduct a visualization study against three representative incomplete multi-modal methods, RFNet [7], mmFormer [8] and IM-Fuse[23], as shown in Figure 10. On the left, we illustrate seven representative missing-modality input combinations, and on the right, we compare the corresponding predictions. With only a single T2 input, our MGML successfully delineates the approximate ET region within the tumor core (TC) while competing methods fail to capture it. With the T1+T2 combination, MGML correctly separates two distinct ET subregions, whereas RFNet, mmFormer, and IM-Fuse tend to merge them into one connected area. When all four modalities are available, MGML maintains accurate boundary delineation and avoids the false-positive regions observed in other methods. These qualitative results, together with the quantitative evidence in Table 3, clearly demonstrate the robustness, adaptability, and generalization ability of our proposed MGML framework under diverse missing-modality conditions.

## 5.3. Ablation Study

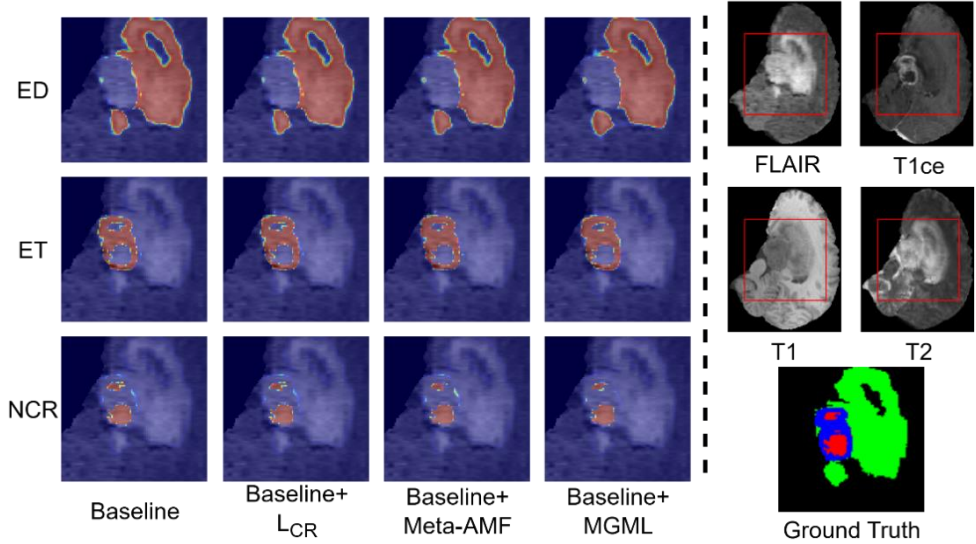

Figure 11. Heatmap Visualization of the Ablation Study for MGML on the Baseline. The left panel displays heatmaps for the ED (edema), ET (enhancing tumor, the contrast-enhanced core on T1ce), and NCR (necrotic or non-enhancing tumor core) categories.

To better understand the role of each component in our framework, we visualize the prediction confidence maps generated by different model variants, including the baseline, baseline+$\mathcal{L}_{CR}$, baseline+ Meta-AMF, and baseline+ MGML. As shown in Figure 11, the baseline model exhibits limited confidence consistency across modalities, especially in regions with ambiguous tumor boundaries, often leading to fragmented or noisy activations. By incorporating the consistency-regularized self-supervised loss ($\mathcal{L}_{CR}$), the model demonstrates improved stability in prediction confidence. The heatmaps become smoother, and spurious activations in irrelevant areas are suppressed. This indicates that the consistency constraint enables the network to better approximate the outputs under complete modality conditions, thus enhancing robustness when some modalities are missing. When applying Meta-AMF alone, the confidence maps highlight more discriminative regions, particularly within tumor cores and enhancing subregions. The adaptive fusion mechanism allows the model to dynamically weigh high-confidence and conservative predictions, leading to sharper boundaries and more reliable confidence distributions. Compared with baseline and baseline+$\mathcal{L}_{CR}$, Meta-AMF introduces a more explicit supervisory signal that effectively guides the model toward modality-aware decision-making. Finally, the full MGML, which integrates both Meta-AMF and $\mathcal{L}_{CR}$, produces the most interpretable confidence maps. The heatmaps reveal both stability and fine-grained precision. Uncertain regions are suppressed through the consistency constraint, while salient structures are enhanced by the meta-guided fusion. This complementary behavior illustrates that our framework explicitly and implicitly promotes multimodal information fusion, leading to higher quantitative performance and more interpretable prediction confidence maps.

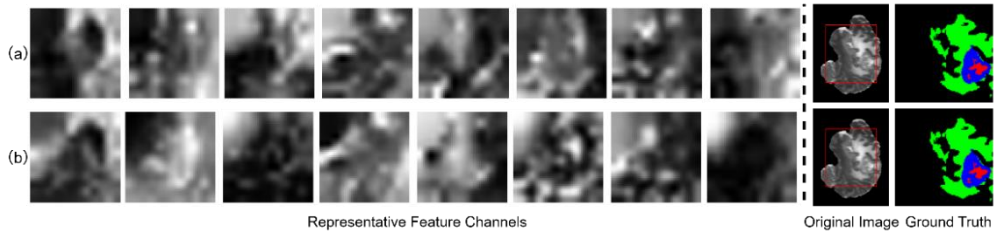

Figure 12. Visualization of representative high-dimensional feature channels from an intermediate layer. The first row (a) corresponds to the baseline model without MGML, while the second row (b) corresponds to the model with MGML integrated.

To provide an intuitive insight into the learned high-dimensional features of our network, several representative feature channels from an intermediate layer before the MGML framework are visualized in Figure 12. Each column represents one feature channel, showing a single middle slice along the D dimension of the 3D feature map. Interestingly, in the baseline model, the feature maps exhibit dispersed activations across the tumor regions, with some channels highlighting partial areas of the tumor. After integrating the MGML framework, the feature maps demonstrate more concentrated and structured activations, particularly around the tumor boundaries, indicating that the module effectively guides the network to focus on the most relevant regions. Moreover, the relative intensity and spatial patterns across slices within each channel suggest that the network captures richer 3D contextual information after the module is applied. These observations validate that the MGML framework enhances the network's ability to integrate multi-modal information, leading to more discriminative feature representations for missing-modality brain tumor segmentation.

### 5.4. Efficiency and Scalability

Table 7. Comparison of baseline models (RFNet, mmFormer and IM-Fuse) with and without MGML in terms of number of parameters, model size, and average training time per epoch.

| Methods | Number of Parameters (M) | Model Size (MB) | Avg Time / Epoch (s) |
| --- | --- | --- | --- |
| RFNet[7] | ~35 | 137 | 65.958±0.203 |
| mmFormer[8] | ~145 | 556 | 41.353±1.089 |
| IM-Fuse[23] | ~254 | 970 | 70.812± 2.154 |
| RFNet[7]+MGML | 35+0.001 | 137+0.005 | 77.331±1.298 |
| mmFormer[8]+MGML | 145+0.001 | 556+0.005 | 44.539±1.741 |
| IM-Fuse[23]+MGML | 254+0.001 | 970+0.005 | 76.587± 2.078 |

As previously stated, our method is only introduced during the training process and does not alter the original architecture of the baseline models. The additional parameters introduced by MGML are almost negligible compared with the massive baselines. As shown in Table 7, for RFNet [7], mmFormer [8] and IM-Fuse[23], the extra parameters and model size brought by MGML are only about +0.001M and +0.005MB, respectively. This confirms that the overhead of our proposed module is minimal. Our MGML framework only adds a few fully connected layers within the MetaNetwork to map input feature maps to meta-parameters (Figure 3). Furthermore, these parameters are optimized only during training based on the soft-label loss and are not saved with the baseline model, which further reduces storage overhead. The extra computational load from backpropagation within the MetaNetwork is also very small.

Beyond parameter and storage analysis, we further evaluate the performance improvement brought by MGML across different architectures. As shown in Table 3, we extended our experiments on three representative SOTA architectures—RFNet, mmFormer, and IM-Fuse, which respectively combine CNN-, Transformer-, and Mamba-based designs. Across these diverse frameworks, MGML consistently improves performance by 0.52%, 1.87%, and 1.18% on RFNet; 3.39%, 2.18%, and 2.36% on mmFormer; and 0.47%, 0.35%, and 0.97% on IM-Fuse in WT, TC, and ET, respectively. These consistent gains across fundamentally different architectures clearly validate that MGML provides a transferable and model-agnostic improvement rather than being a minor modification of existing approaches.

In addition to parameter and storage analysis, we further quantified the training-time overhead introduced by MGML. As shown in Table 7, the average training time per epoch increases moderately when MGML is integrated — by approximately 17.2% for RFNet, 7.7% for mmFormer and 8.1% for IM-Fuse, respectively. This slight increase primarily stems from the dual forward passes of complete and incomplete modalities required for self-distillation. Importantly, this overhead only occurs during training, while the inference cost remains identical to the baseline models since MGML is not used during testing. Given the consistent accuracy gains shown in Table 3, this trade-off between performance improvement and training cost is considered highly acceptable, further demonstrating the practicality of the proposed MGML framework.

Overall, MGML introduces negligible overhead, requires no additional storage, and can be seamlessly integrated into various baseline architectures. This confirms its practicality as a lightweight and flexible training-time module.

**5.5. Limitations and Future Improvements**

While the proposed framework achieves promising performance, several aspects warrant further improvement. The lightweight MetaNetwork may limit its capacity to model complex modality interactions, and the current batch-level parameter generation cannot fully adapt to spatially heterogeneous regions such as tumor boundaries. Future work could explore more expressive or spatially adaptive meta-parameter generation strategies to enhance fusion quality.

Moreover, the smooth approximation functions (smooth max and smooth min) may introduce instability under extreme hyperparameter values, and the random masking strategy might occasionally discard informative areas. Developing confidence-aware masking or normalization strategies could make training more stable and principled.

While MGML has been extensively validated on BraTS2020 and BraTS2023, which already differ in data distribution, acquisition protocols, and annotation standards, these datasets both focus on brain tumor segmentation. In contrast, other public datasets such as ISLES (ischemic stroke lesion segmentation) and ATLAS (stroke lesion segmentation from T1-weighted MRI) address distinct clinical tasks. Moreover, ATLAS provides only single-modality images, and the modalities in ISLES (e.g., DWI, ADC) differ substantially from those used in BraTS. Therefore, applying MGML directly to these datasets falls beyond the current study's scope. Nonetheless, since MGML performs meta-guided adaptive fusion based on available modality signals rather than predefined modality-specific priors, it has strong potential to generalize to other multimodal medical imaging domains. In future work, we plan to extend MGML to additional clinical applications such as stroke lesion analysis and cross-dataset validation to further evaluate its adaptability and robustness across diverse imaging conditions.

Finally, from a clinical perspective, MGML is particularly relevant to clinical scenarios where incomplete or inconsistent MRI contrasts frequently occur due to scanner protocols, time constraints, or patient intolerance. Unlike conventional multimodal methods that require all modalities to be present, MGML's modality-agnostic and meta-learned design allows it to infer optimal fusion strategies from whatever subset of modalities is available. For example, in hospital data where T1c is occasionally missing, the meta-network can adaptively reweight the remaining modalities (e.g., T2/FLAIR) to maintain reliable tumor delineation. This flexibility makes MGML well-suited for real clinical deployment, where data completeness cannot be guaranteed. In future work, we plan to validate MGML on retrospective in-house datasets with naturally missing contrasts to further assess its robustness under practical conditions.

Overall, addressing these limitations will help improve the robustness, stability, and clinical readiness of the proposed MGML framework.

## 6. Conclusion

In conclusion, we propose a lightweight and flexible meta-guided multi-modal learning (MGML) framework to address missing modalities in medical image segmentation. By combining the Meta-AMF module for explicit fusion and the consistency regularization module for implicit fusion, MGML effectively integrates multi-modal information and enhances robustness against missing data. It can be seamlessly embedded into existing architectures without modification, enabling end-to-end learning. Experiments on 15 missing modality combinations demonstrate that MGML achieves superior performance over recent SOTA methods, with average Dice scores of 87.55 (WT), 79.36 (TC), and 62.67 (ET).

**Declaration of competing interest**

The authors declare that there are no conflicts of interest regarding the publication of this paper.

**CRediT authorship contribution statement**

**Yulong Zou**: Conceptualization; Data curation; Formal analysis; Investigation; Methodology; Software; Validation; Visualization; Writing-original draft. **Bo Liu**: Investigation; Validation; Visualization; Writing-review & editing. **Cun-Jing Zheng**: Investigation; Validation; Writing-review & editing. **Yuan-ming Geng**: Investigation; Validation; Writing-review & editing. **Siyue Li**: Validation; Writing-review & editing. **Qiankun Zuo**: Methodology; Resources; Conceptualization; Writing-review & editing. **Shuihua Wang**: Validation, Writing-review & editing. **Yudong Zhang**: Validation, Writing-review & editing. **Jin Hong**: Conceptualization; Data curation; Investigation; Methodology; Resources; Funding acquisition; Project administration; Supervision; Writing-review & editing.

**Acknowledgements**

This work was supported in part by the National Natural Science Foundation of China (62466033), and in part by the Jiangxi Provincial Natural Science Foundation (20242BAB20070).